\documentclass[10pt,journal,compsoc]{IEEEtran}
\pdfoutput=1
\ifCLASSOPTIONcompsoc
  \usepackage[nocompress]{cite}
\else
  \usepackage{cite}
\fi

\usepackage{multirow}
\usepackage{amssymb}
\usepackage{graphicx}
\usepackage{textcomp}
\usepackage{stfloats}
\usepackage{url}
\usepackage{verbatim}
\usepackage{amsmath,amsfonts}
\usepackage{algorithmic}
\usepackage{algorithm}
\usepackage{array}
\usepackage{upgreek}

\usepackage{hyperref}       
\usepackage{microtype}      
\usepackage{xcolor}         

\usepackage{amsmath,amssymb,amsfonts}
\usepackage{float}
\usepackage{subfigure}
\usepackage{makecell}
\usepackage{bm}
\usepackage{wrapfig}
\usepackage{epstopdf}

\ifCLASSINFOpdf
\else
\fi
\begin{document}
\title{
Unifying Graph Contrastive Learning via Graph Message Augmentation }

\author{Ziyan~Zhang, Bo~Jiang*\thanks{* Corresponding author (Email: jiangbo@ahu.edu.cn)}, Jin~Tang and Bin~Luo
\thanks{Ziyan Zhang, Bo Jiang, Jin Tang and Bin Luo are with the School of Computer Science and Technology,  Anhui University, Hefei 230009, China}}

\markboth{Journal of \LaTeX\ Class Files,~Vol.~14, No.~8, August~2015}%
{Shell \MakeLowercase{\textit{et al.}}: Bare Demo of IEEEtran.cls for Computer Society Journals}

\IEEEtitleabstractindextext{%
\begin{abstract}

Graph contrastive learning 
is usually performed by 
first conducting Graph Data Augmentation (GDA) and then employing a contrastive learning pipeline to train GNNs. As we know that GDA is an important issue for graph contrastive learning. Various GDAs have been developed recently which mainly involve dropping or perturbing edges, nodes, node attributes and edge attributes. 
However, to our knowledge, it still lacks a universal and effective augmentor that is suitable for different types of graph data. 
To address this issue, in this paper, 
we first introduce the graph message representation of graph data. 
Based on it, we then propose a novel Graph Message Augmentation (GMA), a \textbf{universal} scheme for reformulating many existing GDAs. 
The proposed unified GMA not only gives a new perspective to understand many existing GDAs but also provides a universal and more effective graph data augmentation for graph self-supervised learning tasks. 
Moreover, GMA introduces an easy way to implement  
the mixup augmentor which is natural for images but usually challengeable for graphs.  
Based on the proposed GMA, we then propose a unified graph contrastive learning, termed Graph Message Contrastive Learning (GMCL), that employs attribution-guided universal GMA  for graph contrastive learning. 
Experiments on many graph learning tasks demonstrate the effectiveness and benefits of the proposed GMA and GMCL approaches. 
\end{abstract}

}

\maketitle

\IEEEdisplaynontitleabstractindextext

%
\IEEEpeerreviewmaketitle

\IEEEraisesectionheading{\section{Introduction}\label{sec:introduction}}

\IEEEPARstart{T}{raditional} 
supervised learning usually relies on large annotated data, resulting in inferior performance in situations where labels are scarce or noisy. 
To address this challenge, 
as a kind of self-supervised learning, contrastive learning~\cite{clsurvey1,clsurvey2} has emerged as an important learning method to learn rich graph representation from unlabeled data. 
It has been extensively employed in various areas such as computer vision~\cite{tpami5}, speech recognition~\cite{clspeech1}, natural language processing~\cite{clnlp1} and graph learning~\cite{zhu2021an,tpami4}.

The typical process for Graph Contrastive Learning (GCL) is performed by first conducting Graph Data Augmentation (GDA) and then employing a contrastive learning pipeline to train graph neural networks such as GCNs~\cite{gcn,tpamiwbb1}, GATs~\cite{gat,tpamiwbb2}, graph Transformers~\cite{chen2022structure,tpami8}, etc. 
In the past years, various GDAs have been developed. One main kind of GDA is to conduct dropping/perturbation operation on node set~\cite{li2021disentangled, graphcl}, edge set~\cite{graphcl, zhu2020deep, zhang2021canonical}, node attributes~\cite{zhang2021canonical, hu2020strategies} 
and edge attributes~\cite{hu2020strategies, NEURIPS2021_85267d34}. 
For example, 
Li et al.~\cite{li2021disentangled} propose three kinds of dropping-based GDA methods including node dropping, edge dropping and node attribute dropping.
Hu et al.~\cite{hu2020strategies} develop a GDA method by masking some elements in both node and edge attributes. 
Manessi et al.~\cite{manessi2021graph} propose a self-supervised auxiliary task which involves perturbation and reconstruction of node and edge attributes. 
Also, inspired by image data augmentation, mixup operation has been employed for GDA. 
However, since it is not straightforward to conduct mixup operation on graphs~\cite{gmixup,wang2021mixup}, it is usually implemented in the semantic-level representation and graphon of different classes. 
For example, Wang et al.~\cite{wang2021mixup} introduce a 
mixup method for different node pairs by mixing the features of receptive field subgraphs.
On the other hand, early GDA methods usually employ random sampling for GDA, leading to the missing of label-invariant information and also introducing noises into GCL learning process. 
To overcome this issue,  recent works have been developed to design some effective learnable graph augmentors~\cite{autoGCL, ADGCL, GCA, you2022bringing}. 
\begin{figure}[!htpb]
\centering
\includegraphics[width=0.48\textwidth]{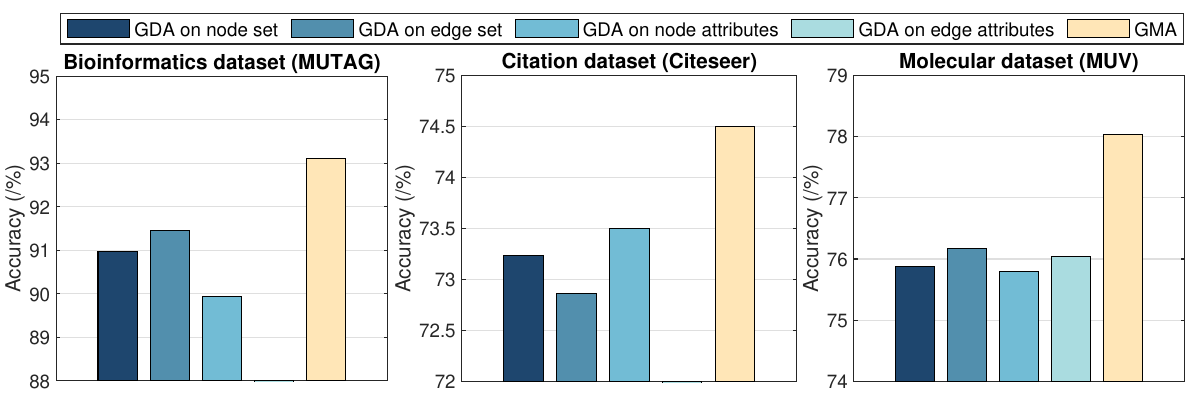}
  \caption{Some comparison performance of different GDAs and our GMA on bioinformatics (MUTAG), citation (Citeseer) and molecular (MUV) datasets. GMA performs consistently better than traditional GDAs.}
\label{fig:GDAGMA}
\end{figure}
\begin{figure*}[!htpb]
\centering
\includegraphics[width=1.0\textwidth]{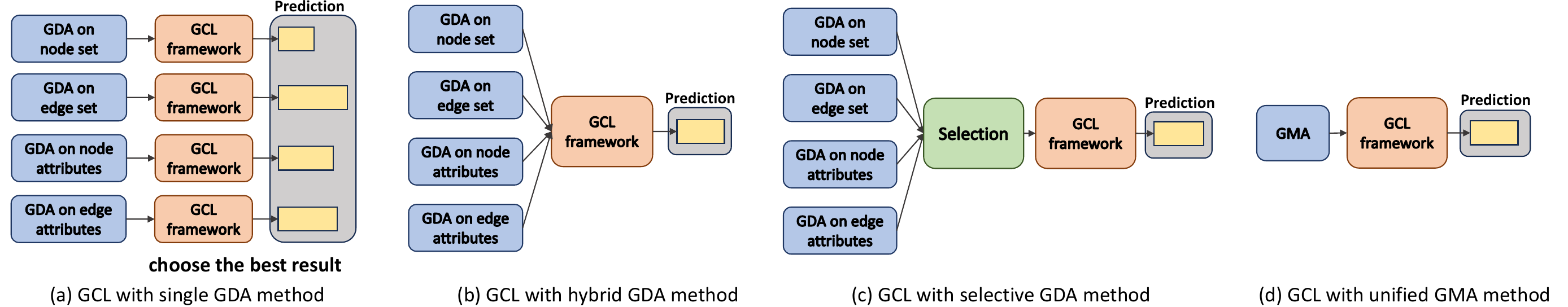}
  \caption{Overview of three kinds of regular GCL frameworks and the proposed GMCL framework.
  }
\label{fig:GDAselect}
\end{figure*}

After reviewing existing works, we observe that 
it is usually necessary to employ different GDAs for different graph data learning scenarios which generally lacks a universal method that is suitable for various graph learning scenarios and tasks. 
Fig.~\ref{fig:GDAGMA} demonstrates some examples which show the comparison results of various GDAs on three different graph datasets.  For more comprehensive comparison results, please refer to $\S$ 6.4. 
Here, we can observe that different graph datasets usually need different GDAs to obtain better performance. 
Therefore, how to determine the most suitable GDA for the specific graph data learning is a problem needing to be studied.   
To address it, 
one simple way is to train and evaluate all candidate GDA methods and select the best GDA method, as shown in Fig.~\ref{fig:GDAselect} (a). 
Obviously, this way requires training multiple GCL models, needing more computational complexity. Also, some works utilize a hybrid way by using multiple GDA modules simultaneously in a single end-to-end GCL framework, as illustrated in Fig.~\ref{fig:GDAselect} (b).
For instance, subgraph sampling~\cite{graphcl,hu2020strategies,jiao2020sub} represents a typical hybrid GDA method, allowing the simultaneous augmentation of node and edge sets. 
GRACE~\cite{zhu2020deep}, CCA~\cite{zhang2021canonical} and GCA~\cite{GCA}  simultaneously apply node attribute masking and edge dropping in their contrastive learning process. 
Another approach is to design a specific GDA selection module to adaptively select the suitable GDA for the downstream task, as illustrated in Fig.~\ref{fig:GDAselect} (c). 
For example,
JOAO~\cite{JOAO} adopts a joint augmentation optimization method to choose the best GDA strategy automatically.
In GraphAug~\cite{luo2023automated}, it utilizes an automatical data augmentation method based on reinforcement learning.
Obviously, the additional GDA selection operation in these methods increases the model parameters and computational complexity. 
After reviewing previous GCLs with various GDAs, it is natural to raise a question: \textit{can we design a universal GDA method that is consistently suitable for various graph learning scenarios ?} 

To address this problem, in this paper, we first propose a novel Graph Message Augmentation (GMA)  by leveraging graph message representation model~\cite{mpnn,dropmessage}. 
The core idea of GMA is conducting various augmentation operations, such as dropping, perturbation and mixup on graph messages rather than on the original graph directly.  
The main advantages of  GMA are three aspects. 
\textbf{First}, 
GMA provides a {universal} data augmentor for graphs. It integrates the cues of graph nodes and edges together in a unified formulation for graph data augmentation. 
Also, GMA is effective and performs \emph{consistently} better than traditional GDAs, as shown in Fig.~\ref{fig:GDAGMA} and further validated in Experiments. 
\textbf{Second}, 
GMA can unify many traditional GDA methods including dropping/perturbation on graph node set, edge set, node attributes and edge attributes, etc. 
We can prove that many traditional GDAs can be regarded as the special cases of GMA, as presented in $\S$ 4. 
This provides a unified understanding for traditional GDAs. 
\textbf{Third}, GMA provides an easy and natural way to implement the mixup augmentor which is an important data augmentation strategy and typically challenging to be defined on graphs~\cite{gmixup}.   
%
%

Based on  GMA, we then introduce a general graph contrastive learning scheme, termed  Graph Message Contrastive Learning (GMCL), for graph self-supervised learning. 
The overall architecture of GMCL is shown in Fig.~\ref{fig:GDAselect} (d). 
GMCL can be implemented in either random or adaptive manner. 
For the adaptive GMCL way, we design a novel Attribution-guided Graph Message Augmentor (AttGMA) module which can adaptively learn the augmentation indicators to preserve the label-invariant information in graph augmentation. 

In summary, this paper presents the following main contributions:
\begin{itemize}
   \item We propose a universal Graph Message Augmentation (GMA) by conducting dropping, perturbation and mixup on graph message representation. GMA can unify various traditional GDA methods and provides a unified understanding for them. 
 
   \item Based on graph message representation, we derive a new simple and effective mixup augmentor, i.e., graph message mixup, for graph data augmentation and graph  contrastive learning task.

   \item Based on GMA, we present a simple and unified contrastive learning architecture, termed Graph Message Contrastive Learning (GMCL), for graph data self-supervised learning. 

   \item We design a new learnable Attribution-guided Graph Message Augmentor (AttGMA) module for adaptive graph message augmentation to preserve the label-invariant information of augmented data. 
 
\end{itemize}
Experimental results on four types of graph datasets demonstrate the effectiveness and benefits of the proposed GMCL approach on various graph learning tasks. 
The remainder of this paper is organized as follows. 
In $\S$ 2, we revisit some related works on graph contrastive learning and graph data augmentation. 
In $\S$ 3, we review several commonly used GDA techniques in contrastive learning.
In $\S$ 4, we present our graph message augmentation scheme. 
In $\S$ 5, we introduce the proposed graph message contrastive learning network. 
We evaluate the proposed methods on various graph learning tasks in $\S$ 6. 

\section{Related work}

\textbf{Graph Contrastive Learning. }
Graph Contrastive Learning (GCL) based self-supervised training provides a powerful method to alleviate the growing demand for labeled data on training graph neural networks.
From the perspective of contrastive loss, GCLs can be divided into same-scale contrast methods and cross-scale contrast methods.
For the same-scale contrast methods, 
for example, 
GCC~\cite{GCA}, GRAND~\cite{grand} and CCA~\cite{zhang2021canonical} focus on node-level representation tasks and conduct node-level contrastive learning.
These methods generally first consider the same nodes in different augmented graphs as positive pairs and then maximize the similarities between  positive pairs. 
In addition, some other methods such as GraphCL~\cite{graphcl}, InforBN~\cite{infograph}, AD-GCL~\cite{ADGCL} and AutoGCL~\cite{autoGCL} propose to utilize global contrastive learning for graph-level representation tasks. 
These methods typically strive to maximize the mutual information between graph-level representations which are generated from various augmentors. 
For cross-scale contrast methods, they generally utilize a cross-scale contrastive loss by contrasting local and global representations. 
For example, DGI~\cite{DGI} and its derivatives~\cite{HDGI,STDGI} propose to maximize the mutual information between global graph-level representation and node representations. 
GraphLoG~\cite{xu2021self}, PGCL~\cite{lin2022prototypical} and GPCL~\cite{peng2022graph} propose to employ prototype-based contrastive learning methods, aiming to contrast local instances and global prototype representations. 

\textbf{Graph Data Augmentation. }
One important aspect of GCL is to design various Graph Data Augmentations (GDAs) for assisting GNN backbone to learn the inherent invariant representations for graphs. 
GDAs are usually implemented  
by utilizing dropping or perturbation operation on node set, edge set, node attributes and edge attributes. 
For example,
GraphCL~\cite{graphcl}, AD-GCL~\cite{ADGCL} and AutoGCL~\cite{autoGCL} generate the augmented graph data by dropping some nodes or edges randomly. 
GCA~\cite{GCA}, CCA~\cite{zhang2021canonical} and GRAND~\cite{grand} mask some node/edge attributes to obtain the new augmented graph data. 
MGAE~\cite{MGAE} and Corrupted Features Reconstruction~\cite{manessi2021graph} propose to perturb node and edge features with random noises.  
In addition, some other methods, such as graph diffusion~\cite{hassani2020contrastive}, constructing hypergraph~\cite{xia2021self} and graphon mixup~\cite{gmixup}, have also been proposed in recent years. 

Many of the above GDAs~\cite{graphcl,grand,MGAE,manessi2021graph,tpami3} generate various augmented data in random manners. 
However, the random-based augmentations might violate the assumption of label invariance~\cite{autoGCL,ADGCL}.
Some recent works are devoted to overcome this issue by designing the learnable graph augmentors. 
For example, 
Yin et al.~\cite{autoGCL} propose a learnable view generator which is optimized by minimizing both similarity and contrastive loss functions. 
You et al.~\cite{you2022bringing} employ graph generative model to generate a learnable prior and regularize the generative model by leveraging InfoMin~\cite{tian2020makes} and InfoBN~\cite{alemi2016deep} principles. 
Suresh et al.~\cite{ADGCL} utilize an adversarial network which encourages the GDA module to focus on the label-related information and remove some redundant information. 
Zhao et al.~\cite{zhao2021data} propose GAug to learn a learnable edge probability matrix for GDA to promote GCL performance. 

Considering the heterogeneity of practical graph data, 
single GDA methods~\cite{graphcl,grand,autoGCL,ADGCL} 
are usually limited in broader applicability. 
To address this limitation, 
some GCL frameworks propose to employ the hybrid GDA strategies, i.e., using multiple GDA methods simultaneously. 
For example, 
subgraph sampling~\cite{graphcl,hassani2020contrastive,hu2020strategies} represents a typical hybrid GDA method, allowing the simultaneous augmentation of node and edge sets. 
GRACE~\cite{zhu2020deep}, CCA~\cite{zhang2021canonical} and GCA~\cite{GCA}  simultaneously apply node attribute masking and edge dropping in their contrastive learning process. 
In addition, some other recent GCLs employ selective GDA methods to adaptively select the suitable GDA for the downstream tasks. 
For example,
You et al.~\cite{JOAO} propose Joint Augmentation Optimization (JOAO) strategy to choose an adaptive GDA  automatically. 
Zhao et al.~\cite{zhao2022autogda} formulate the GDA selection problem as a bi-level optimization problem and propose a reinforcement learning framework AutoGDA to address it. 
Luo et al.~\cite{luo2023automated} propose GraphAug which utilizes a trainable model to automatically select the best GDA.


\section{Graph Data Augmentation Revisited} 

\begin{figure*}[!htpb]
\centering
\includegraphics[width=1.0\textwidth]{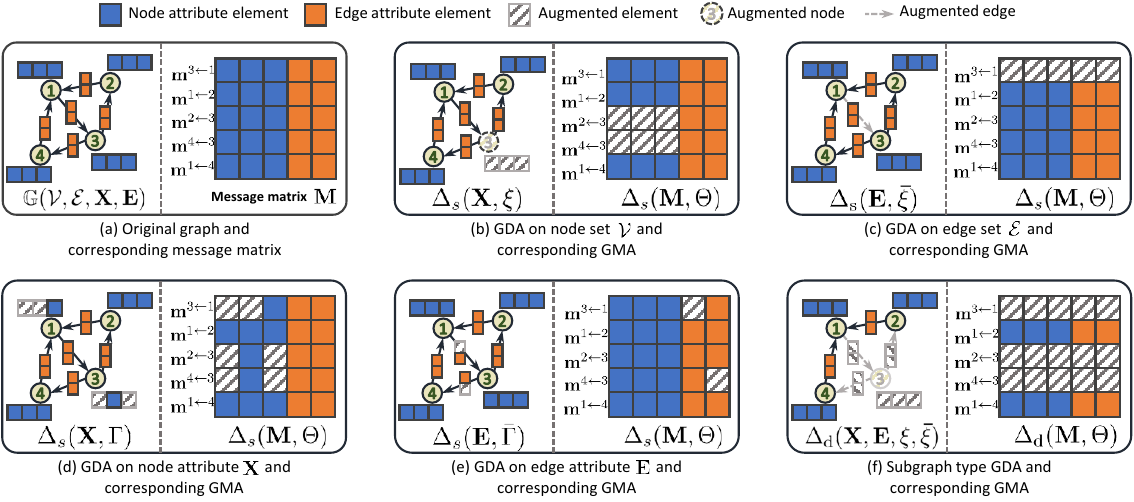}
  \caption{(a) An example of the graph with its message representation Eq.(9). (b-f) Demonstration of different GDAs and corresponding GMA forms.}
\label{fig:demo}
\end{figure*}

It is known that Graph Data Augmentations (GDAs) have been commonly employed in graph contrastive learning. 
In this section, we review several commonly used GDA techniques in graph contrastive learning tasks. 
Given an attributed relational graph $\mathbb{G}(\mathcal{V},\mathcal{E}, \mathbf{X}, \mathbf{E})$,
where $\mathcal{V}$ and $\mathcal{E}$ denote the node and edge sets respectively,  
each node $v$ has a $d$-dimension feature vector $\mathbf{x}_v$ and $\mathbf{X}=\{\mathbf{x}_1,\mathbf{x}_2\dots\mathbf{x}_{|\mathcal{V}|}\}\in\mathbb{R}^{|\mathcal{V}|\times d}$ denotes the collection of node attributes. 
Each edge $e_{vh}$ has a $p$-dimension feature vector and $\mathbf{E}=\{\mathbf{e}_{vh}\}\in\mathbb{R}^{|\mathcal{E}|\times p}$ denotes the collection of edge features. 
GDA aims to generate a new graph $\mathbb{G}'$ by typically performing dropping, perturbation or mixup operation on $\mathbb{G}$. 
Usually, the dropping and perturbation operation are conducted on node set $\mathcal{V}$, edge set $\mathcal{E}$, node attributes $\mathbf{X}$ and edge attributes $\mathbf{E}$~\cite{graphcl, NEURIPS2021_85267d34, MGAE},
as illustrated in Fig.~\ref{fig:demo}. The details are introduced below. 

\emph{(a) GDA on node set $\mathcal{V}$.}
Let $\mathbf{\xi}\in \{0,1\}^{|\mathcal{V}|}$ denote the indicator vector of node augmentation, where $\mathbf{\xi}_h=1$ indicates that the $h$-th node is dropped or perturbed, i.e., the $h$-th row of feature matrix $\mathbf{X}$ is set to $\mathbf{0}$ or perturbed randomly. The node set type GDA can be formulated as, 
\begin{equation}\label{EQ:GDAN}
\mathbf{X}'=\Delta_{\mathrm{s}} (\mathbf{X},\mathbf{\xi}) 
\end{equation}
where $\mathrm{s}\in\{\mathrm{d}, \mathrm{p}\}$ denotes the dropping or perturbation operation.

\emph{(b) GDA on edge set $\mathcal{E}$.}
Let $\mathbf{\bar{\xi}}\in \{0,1\}^{|\mathcal{E}|}$ denote the indicator vector of edge augmentation, where $\mathbf{\bar{\xi}}_{e_{vh}}=1$ indicates that the edge $e_{vh}$ is dropped or perturbed, i.e., feature vector $\mathbf{e}_{vh}$ in $\mathbf{E}$ is set to $\mathbf{0}$ or perturbed randomly. 
The edge set type GDA can be formulated as,  
\begin{equation}\label{EQ:GDAE}
\mathbf{E}'=\Delta_{\mathrm{s}} (\mathbf{E}, \mathbf{\bar{\xi}})
\end{equation}

\emph{(c) GDA on node attributes $\mathbf{X}$.}
Let ${\Gamma}\in \{0,1\}^{|\mathcal{V}|\times d}$ denote the indicator matrix of node attributes, where ${\Gamma}_{v,k}=1$ indicates that element $\mathbf{X}_{v,k}$ is dropped or perturbed, i.e., $\mathbf{X}_{v,k}$ is set to ${0}$ or perturbed randomly. The node attribute type GDA can be formulated as, 
\begin{equation}\label{EQ:GDANF}
\mathbf{X}'=\Delta_{\mathrm{s}} (\mathbf{X}, {\Gamma})
\end{equation}

\emph{(d) GDA on edge attributes $\mathbf{E}$.}
Let ${\bar{\Gamma}}\in \{0,1\}^{|\mathcal{E}|\times p}$ denote the indicator matrix of edge attributes, where ${\bar{\Gamma}}_{e_{vh},k}=1$ indicates that element $\mathbf{E}_{e_{vh},k}$ is dropped or perturbed, i.e., the element $\mathbf{E}_{e_{vh},k}$ is  set to ${0}$ or perturbed randomly. The edge attribute GDA can be formulated as, 
\begin{equation}\label{EQ:GDAEF}
\mathbf{E}'=\Delta_{\mathrm{s}} (\mathbf{E}, {\bar{\Gamma}})
\end{equation}

\emph{(e) Subgraph type GDA.}
Subgraph GDA is conducted by considering both node and edge set dropping.  
Let $\mathbf{\xi}\in \{0,1\}^{|\mathcal{V}|}$ and $\mathbf{\bar{\xi}}\in \{0,1\}^{|\mathcal{E}|}$ denote the dropping indicator vector for node and edge respectively,
where $\mathbf{\xi}$ and $\mathbf{\bar{\xi}}$ are defined as same as the node set type and edge set type GDA. 
The subgraph type GDA can be formulated as, 
\begin{equation}\label{EQ:GDAS}
\{\mathbf{X}',\mathbf{E}'\} = \Delta_{\mathrm{d}} (\mathbf{X}, \mathbf{E}, \mathbf{\xi},\mathbf{\bar{\xi}})
\end{equation}

\emph{(f) Mixup type GDA.} 
The aim of mixup
is to produce a new augmented graph by mixing the information from several different graphs. 
Mixup operation has been usually conducted on grid type data such as images for data augmentation. 
It is generally not straightforward to achieve mixup operation on graphs due to the different structures of graphs. 
Previous works propose to conduct mixup operation on graphs by 
transforming graphs into semantic-level representations~\cite{wang2021mixup} or graphons~\cite{gmixup} with different classes. 

\section{Graph message augmentation}

\subsection{Graph Message Representation}

Given a directed~\footnote{Here, we focus on directed graph. For undirected graph, it can be equivalently transferred to a bi-directed graph. Thus, the introduced methods are suitable for both directed and undirected graphs. } attributed relational graph $\mathbb{G}(\mathcal{V},\mathcal{E}, \mathbf{X}, \mathbf{E})$,
where $\mathcal{V}$ and $\mathcal{E}$ denote the node and (directed) edge set respectively,  
where $\mathbf{X}=\{\mathbf{x}_1\dots\mathbf{x}_{|\mathcal{V}|}\}\in\mathbb{R}^{|\mathcal{V}|\times d}$ denotes the collection of node attributes and $\mathbf{E}=\{\mathbf{e}_{vh}\}\in\mathbb{R}^{|\mathcal{E}|\times p}$ denotes the collection of edge features,  
it is known that the layer-wise propagation of GNNs~\cite{gcn,gat,gin} involves two phases, i.e., message aggregation and feature transformation (embedding). 
To be specific, we can generally formulate GNN's layer-wise propagation as
\begin{align}
\mathbf{z}_v &= \sum_{h\in \mathcal{N}_v} \mathcal{M}(\mathbf{x}_h, \mathbf{e}_{vh})\label{EQ:GNN1}\\ 
\tilde{\mathbf{x}}_v &= \mathcal{F}(\mathbf{z}_v,\mathbf{W}) \label{EQ:GNN2}
\end{align}
%
%
%
where $\mathcal{N}_v$ denotes the neighboring set of node $v$ including node $v$ itself.  
$\mathcal{M}(\cdot)$ represents the message representation function and 
$\mathcal{F}(\cdot)$ denotes the feature transformation function such as linear projection, MLP, etc.  
For example, Kipf et al.~\cite{gcn} propose Graph Convolution Network (GCN) which can be regarded as defining $\mathcal{M}(\cdot)$ as
\begin{align}
\mathcal{M}(\mathbf{x}_h, \mathbf{e}_{vh}) = ({\mathbf{D}^{-\frac{1}{2}}(\mathbf{I}+\mathbf{A})\mathbf{D}^{-\frac{1}{2}}})_{v,h} \mathbf{x}_h
\label{EQ:GCN}
\end{align}
where ${\mathbf{A}}$ denotes the adjacency matrix. 
Obviously, Eq.(\ref{EQ:GCN}) is only suitable for the weighted graphs. For the edges with high-dimensional features, Eq.(\ref{EQ:GCN}) is not suitable. 
Duvenaud et al.~\cite{duvenaud2015convolutional}  propose the layer-wise propagation which can be regarded as defining the message function  as
\begin{align}
\mathcal{M}(\mathbf{x}_h, \mathbf{e}_{vh}) = \big[\mathbf{x}_h \| \mathbf{e}_{vh}\big]\label{EQ:MGCN}
\end{align}
where $\|$ denotes the concatenation operation. 

In this paper, we mainly focus on message function $\mathcal{M}(\cdot)$. 
Similar to work~\cite{dropmessage}, we define \textbf{graph message} $\mathbf{m}^{v\leftarrow h}$ propagated from neighboring node $h$ to center node $v$ as 
\begin{align}\label{EQ:message}
&\mathbf{m}^{v\leftarrow h} = \mathcal{M}(\mathbf{x}_h,\mathbf{e}_{vh})
\end{align}
 where $h\in \mathcal{N}_v$. 
 Note that, one property of graph messages is that they can be well represented as the
matrix form. 
Specifically, we can use $\mathbf{M}=\{\mathbf{m}^{v\leftarrow h}|h\in \mathcal{N}_v, v=1, 2\cdots |\mathcal{V}|\}$
to denote the collection of all messages propagated on the graph, where each row of message matrix $\mathbf{M}$ denotes a specific message propagated on each edge. 
In this paper, we  utilize graph message definition Eq.(9) in introducing our methodology and experiments.

\subsection{Unifying Graph Data Augmentation via Graph Message Augmentation}

Using the above message representation Eq.(\ref{EQ:message}), we can propose a more general graph augmentation strategy and also achieve mixup type graph augmentation more intuitively. We call it as Graph Message Augmentation (GMA). 

\subsubsection{GMA via graph message dropping/perturbation} 
We can randomly or adaptively drop/perturb some elements on graph message ${\mathbf{M}}$ to generate the augmented graph message ${\mathbf{M}'}$. 
Formally, let $\Theta$ denote the 
indicator matrix where $\Theta_{v,h}=1$ indicates that the element $\mathbf{M}_{v,h}$ is dropped/perturbed, i.e., $\mathbf{M}_{v,h}$
is set to 0 or perturbed randomly, 
and $\Theta_{v,h}=0$ otherwise. 
Then, our graph message dropping/perturbation based GMA can be generally formulated as follows, 
\begin{equation}\label{EQ:GMA}
\mathbf{M}'=\Delta_{\mathrm{s}} (\mathbf{M}, {\Theta})
\end{equation}
where $\mathrm{s}\in \{\mathrm{d},\mathrm{p}\}$ denotes the 
dropping or perturbation operation. 
$\Theta$ can be generated randomly or learned adaptively, as presented in $\S$ 5.1 and $\S$ 5.2 respectively. 

\textbf{Remark.} Using graph message definition Eq.(9), the above GMA  can unify various GDA strategies (Eq.(\ref{EQ:GDAN})-Eq.(\ref{EQ:GDAS})) mentioned before, i.e., each GDA can be regarded as a special form of the proposed GMA with graph message definition Eq.(9). 
This can be analyzed as follows. 
Let $\Theta=\{{\uptheta}^{v\leftarrow h}|h\in\mathcal N_v, v=1,2\cdots |\mathcal{V}|\}\in\{0,1\}^{|\mathcal{E}|\times (d+p)}$ 
be the augmentation indicator matrix for the message matrix $\mathbf{M}\in\mathbb{R}^{|\mathcal{E}|\times (d+p)}$, 
where each row   ${\uptheta}^{v\leftarrow h}\in\{0,1\}^{d+p}$ denotes the augmentation indicator vector for the graph message $\mathbf{m}^{v\leftarrow h}$. 
Then, we can have the following,  

{(a) For node set augmentation Eq.(\ref{EQ:GDAN}), it can be equivalently conducted on graph message matrix with the augmentation indicator matrix ${\Theta}$ as
\begin{equation}\label{EQ:thetan}
\uptheta^{v\leftarrow h}_{i} = \left\{
\begin{aligned}
&1 \ \ if \ \ \mathbf{\xi}_h=1  \ \ and \ \ 1\leq i\leq d \\
&0 \ \ otherwise
\end{aligned}
\right.
\end{equation}
where $d$ denotes 
the feature dimension of nodes. 
}

{(b) For edge set augmentation Eq.(\ref{EQ:GDAE}), it can be equivalently conducted on graph message matrix with the augmentation indicator matrix ${\Theta}$ as
\begin{equation}\label{EQ:thetae}
\uptheta^{v\leftarrow h}_i = \left\{
\begin{aligned}
&{1} \ \ if \ \ \mathbf{\bar{\xi}}_{e_{vh}}=1 \ \ \ and \ \ 1\leq i\leq d+p \\
&{0} \ \ otherwise
\end{aligned}
\right.
\end{equation}
where $p$ denotes the feature dimension of edges. 
}

{(c) For node attribute augmentation Eq.(\ref{EQ:GDANF}), it can be equivalently conducted on graph message matrix with the augmentation indicator matrix ${\Theta}$ as
\begin{equation}\label{EQ:thetanf}
\uptheta^{v\leftarrow h}_{i} = \left\{
\begin{aligned}
&1 \ \ if \ \ {\Gamma}_{h,i}=1 \ \ and \ \ 1\leq i\leq d
\ \ \ \\
&0 \ \ otherwise
\end{aligned}
\right.
\end{equation}
}

{(d) For edge attribute augmentation Eq.(\ref{EQ:GDAEF}), it can be equivalently conducted on graph message matrix with the augmentation indicator matrix ${\Theta}$ as 
\begin{equation}\label{EQ:thetaef}
\uptheta^{v\leftarrow h}_{i} = \left\{
\begin{aligned}
&1 \ \ if \ \ \bar{\Gamma}_{e_{vh},i-d}=1 \ \ and \ \ d< i\leq d+p
\ \ \ \\
&0 \ \ otherwise
\end{aligned}
\right.
\end{equation}
}

(e) For subgraph type augmentation Eq.(\ref{EQ:GDAS}), it can be equivalently conducted on graph message matrix with the augmentation indicator matrix ${\Theta}$ as 
\begin{equation}\label{EQ:thetas}
\uptheta^{v\leftarrow h} = \left\{
\begin{aligned}
&\mathbf{1} \ \ if \ \ \mathbf{\xi}_{h}=1 \ \ or \ \ \bar{\mathbf{\xi}}_{e_{vh}}=1 \\
&\mathbf{0} \ \ otherwise
\end{aligned}
\right.
\end{equation}

Fig.~\ref{fig:demo} (b-f) demonstrate the examples to illustrate the correspondence between each GDA and our GMA.
We can note that each kind of GDA can be equivalently reformulated in our GMA. 
This  provides a unified understanding
for traditional GDAs. 

\subsubsection{GMA via graph message mixup} 
As an important augmentor, mixup has been commonly used for grid type data (\emph{e.g.,} images)  augmentation. 
As we know that, it is 
usually not straightforward to 
define mixup operation on graphs~\cite{wu2022graphmixup,gmixup}. 
However, using the above graph message representation, we can easily achieve mixup operation, i.e., graph message mixup, for graph data augmentation. 
Specifically, the aim of the proposed graph message mixup is to produce a new augmented graph message by mixing two messages from two different graphs. 
In this paper, we adopt the label-invariant mixup strategy~\cite{mo2021object} in our message mixup. 

\begin{figure}[!htpb]
\centering
\includegraphics[width=0.48\textwidth]{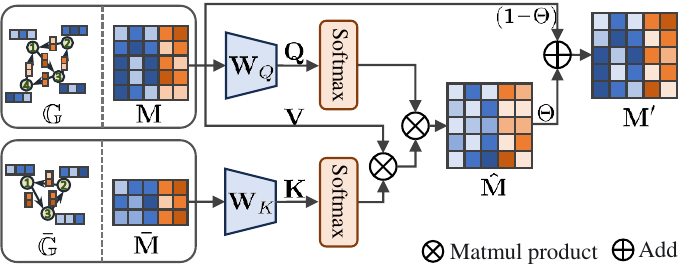}
  \caption{The pipeline of GMA via graph message mixup.}
\label{fig:mixup}
\end{figure}

Given two graphs $\mathbb{G}$ and $\mathbb{\bar{G}}$ with the associated graph message matrix $\mathbf{M}$ and $\mathbf{\bar{M}}$, 
the label-invariant graph message mixup aims to combine the information of both $\mathbf{M}$ and $\mathbf{\bar{M}}$ to generate a new augmentation $\mathbf{M}'$ with the same label as $\mathbf{M}$. 
Since $\mathbf{\bar{M}}$ is usually not the same size with $\mathbf{M}$ due to different sizes of graph $\mathbb{G}$ and $\mathbb{\bar{G}}$, we first obtain the mixed graph message $\mathbf{\hat{M}}$ with the same size with $\mathbf{M}$ via a simple linear cross attention mechanism~\cite{shen2021efficient} as 
\begin{align}
&\mathbf{Q}=\mathbf{M}\mathbf{W}_Q, \mathbf{K}=\mathbf{\bar{M}}\mathbf{W}_K, \mathbf{V}=\mathbf{M}\\
&\mathbf{\hat{M}} = \mathrm{Softmax}(\mathbf{Q})\big(\mathrm{Softmax}(\mathbf{K})^T \mathbf{V}\big)\label{EQ:mixup}
\end{align}
where $\mathbf{W}_Q$ and $\mathbf{W}_K$ are learnable transformation parameters. 
Obviously, $\hat{\mathbf{M}}$ incorporates the information of $\mathbf{M}$ and $\bar{\mathbf{M}}$ while has the same size with $\mathbf{M}$. 
Then, 
we can conduct graph message mixup as follows, 
\begin{align}\label{EQ:gm_mixup}
\mathbf{M}' &= \Delta_m(\mathbf{M}, \mathbf{\bar{M}},\Theta) = (\mathbf{1}-\Theta) \odot \mathbf{M} + \Theta \odot \mathbf{\hat{M}}
\end{align}
where $\Theta_{v,h}\in\{0,1\}$ denotes the indicator matrix to indicate whether $\mathbf{M}_{v,h}$ is replaced with $\mathbf{\hat{M}}_{v,h}$.
Here, $\Theta$ can be generated either randomly or learned adaptively, as introduced below. 
Fig.~\ref{fig:mixup} demonstrates the whole pipeline of graph message mixup.



\section{Graph Message Contrastive Learning}
\begin{figure*}[!htpb]
\centering
\includegraphics[width=0.9\textwidth]{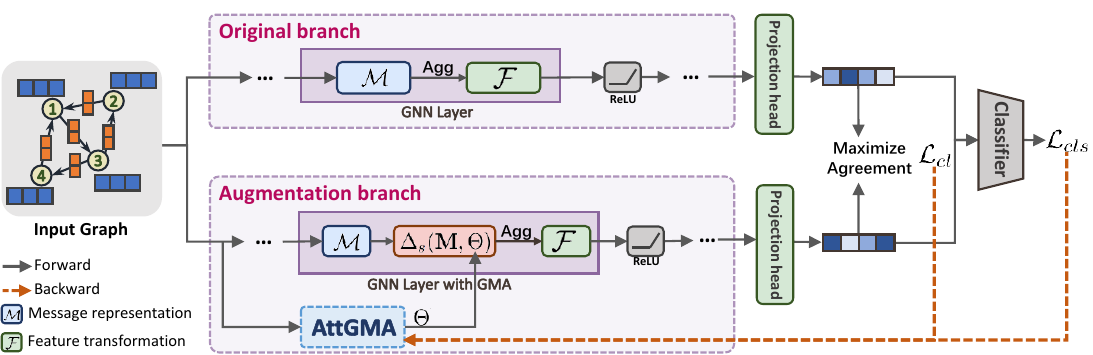}
  \caption{The architecture of Graph Message Contrastive Learning (GMCL) framework for graph classification tasks.}
\label{fig:GMCL}
\end{figure*}


Based on the above proposed GMAs, we present a more general contrastive learning scheme, termed Graph Message Contrastive Learning (GMCL), for graph self-supervised representation. 
Following regular contrastive 
learning architecture~\cite{autoGCL,ADGCL}, 
the proposed GMCL framework generally adopts a two-branch GNN architecture, 
with one branch being the original GNN and the other being GNN with GMA. 
All branches share the learnable network weight parameters (Eq.(\ref{EQ:GNN2})). 
Fig.~\ref{fig:GMCL} shows the architecture of GMCL for the graph classification tasks. It contains GNN branches, projection head and classifier module in which the projection head aims to obtain the global graph representation for   graph classification task, as utilized in works~\cite{DGI,graphcl}. 


\subsection{Random GMA for Contrastive Learning}

Similar to   previous GDA methods~\cite{graphcl,grand},
we can naturally propose a straightforward GMA based on random strategy, i.e., generating indicator matrix $\Theta$ randomly. 
Specifically, the probability of generating each element in $\Theta$ follows a default i.i.d. uniform distribution (or any other distribution).
Then, based on indicator matrix $\Theta$, the new augmentation of graph messages can be generated by using dropping, perturbation and mixup strategies (Eqs.(\ref{EQ:GMA},\ref{EQ:gm_mixup})). 
Finally, we adopt the above GMCL pipeline (Fig.~\ref{fig:GMCL} without AttGMA module) to learn the effective data representation for the input graph.

\subsection{Adaptive GMA for Contrastive Learning}
\begin{figure}[!htpb]
\centering
\includegraphics[width=0.48\textwidth]{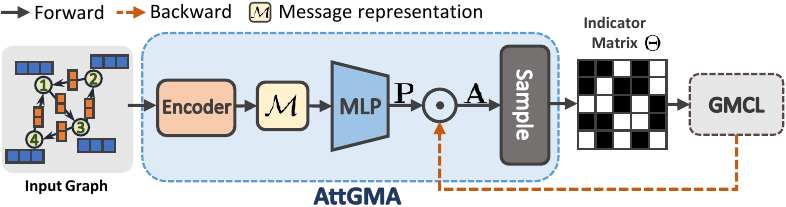}
  \caption{The architecture of Attribution-guided Graph Message Augmentor (AttGMA) module.}
\label{fig:LGMA}
\end{figure}

The above random GMA method may cause the loss of label invariant information in graph messages and introduce the noises into  GMCL training process. 
To overcome this issue, in this section, we further propose an adaptive GMA using a learnable indicator matrix $\Theta$ in the augmented message via an Attribution-guided Graph Message 
 Augmentor (AttGMA) module, as shown in Fig.5.
Specifically, as illustrated in Fig.~\ref{fig:LGMA},  
AttGMA uses an additional Encoder and employs MLP on the extracted messages to learn the probability matrix $\mathbf{P}\in\mathbb{R}^{|\mathcal{E}|\times(d+p)}$ as follows, 
\begin{align}
\label{EQ:LGAM1}
\mathbf{P} = \mathrm{MLP}(\mathcal{M}(\mathrm{Encoder} (\mathbb{G})))
\end{align}
Here, we adopt GIN~\cite{gin} as Encoder module. 
Based on probability matrix $\mathbf{P}$, one simple way is to  generate a learnable indicator matrix via some sampling techniques, as utilized in  works~\cite{autoGCL, ADGCL}. 
However, to enhance the generation of the label-invariant GMA, we further introduce the attribution model~\cite{hao2021self,addrop} which aims to utilize the gradient calculation to reflect the element importance of graph messages for the final learning task. 
We propose to learn  a message attribution matrix $\mathbf{A}\in\mathbb{R}^{|\mathcal{E}|\times(d+p)}$ as 
\begin{align}
\label{EQ:LGAM2}
\mathbf{A} = \frac{\partial \mathcal{L}_a}{\partial \mathbf{P}}
\end{align}
where $\mathcal{L}_a$ denotes the loss computed via Eq.(\ref{EQ:AD_loss}) in $\S$ 5.3. 
Note that, the higher $\mathbf{A}_{e_{vh},k}$ indicates that the message  element $\mathbf{m}^{v\leftarrow h}_{k}$ contributes more to the final learning tasks. 
Based on $\mathbf{A}$, we then generate the binary indicator matrix $\Theta$ by using Gumbel-Softmax~\cite{jang2016categorical,tpami7}  as
\begin{align}
\label{EQ:LGAM3}
\Theta = \mathrm{GumbelSoftmax}(\mathbf{P}\odot\mathbf{A})
\end{align}
where $\odot$ denotes the Hadamard product. 
Fig.~\ref{fig:LGMA} demonstrates the architecture of the proposed AttGMA module.

\textbf{Contrastive Loss.} 
Our GMCL follows the basic contrastive learning loss
which aims to maximize the consistency between positive pairs and minimize the consistency between negative pairs.
Similar to previous works~\cite{graphcl,autoGCL,grand}, the positive pair is defined as the graph/node representation of the original graph and its augmented graph. The negative pairs are defined as the representations of different graphs/nodes.
We use the commonly used normalized temperature-scaled cross entropy loss function~\cite{sohn2016improved, graphcl} as
\begin{equation}
\mathcal{L}_{cl}(\tilde{\mathbf{X}},\tilde{\mathbf{X}}') =-\frac{1}{n}{\rm log}\frac{\sum^{n}_{i=1}{\rm exp}({\rm sim}(\tilde{\mathbf{x}}_i,\tilde{\mathbf{x}}'_i)/\tau)}{\sum_{i,j=1,i\neq j}^{n}{\rm exp}({\rm sim}(\tilde{\mathbf{x}}_i,\tilde{\mathbf{x}}'_j)/\tau)}
\end{equation}
where $\tau$ represents the temperature parameter. $n$ denotes the number of graphs/nodes.
$\tilde{\mathbf{X}}$ denotes the representation output from the original GNN branch
and $\tilde{\mathbf{X}}'$ denotes the output representation  from the augmentation GNN branch. 
$\rm{sim}(\cdot)$ denotes the cosine similarity function.
Also, for semi-supervised tasks, we can further add the cross-entropy classification loss as 
\begin{equation}
\mathcal{L}_{cls}(\mathbf{U},\mathbf{Y}) = -\frac{1}{B}\sum_{i=1}^{B}\sum_{c=1}^{C}\mathbf{Y}_{i,c}{\rm log}(\mathbf{u}_{i,c})
\end{equation}
where $\mathbf{Y}\in\mathbb{R}^{B\times C}$ denotes the ground-truth label matrix for labeled graphs/nodes, where $B$ denotes the number of labeled nodes and $C$ denotes the class number. $\mathbf{U}$ denotes the output of classifier from original GNN branch. 
We also compute the classification loss based on the classifier output $\mathbf{U}'$  for the augmentation branch.
Overall, the total loss is 
\begin{equation}\label{EQ:loss}
\mathcal{L} = \mathcal{L}_{cls}(\mathbf{U},\mathbf{Y})+ \mathcal{L}_{cls}(\mathbf{U}',\mathbf{Y}) + \alpha \mathcal{L}_{cl}(\tilde{\mathbf{X}},\tilde{\mathbf{X}}')
\end{equation}
where $\alpha$ is the hyperparameter balancing contrastive loss and classification loss.
All modules are optimized jointly in an end-to-end manner.

\subsection{Implementation Detail}
The data augmentation module AttGMA is only utilized and optimized during the training process. 
In the testing process, we just use the GNN encoder and classifier to obtain the final representation and label prediction, as commonly conducted in previous works~\cite{autoGCL,ADGCL}. 
To calculate the attribution matrix,
we first perform GMCL by removing $\mathbf{A}$ in Eq.(\ref{EQ:LGAM3}) to obtain the loss $\mathcal{L}_a$ as 
\begin{equation}\label{EQ:AD_loss}
\mathcal{L}_a = \mathcal{L}_{cls}(\mathbf{U}',\hat{\mathbf{Y}}) + \alpha\mathcal{L}_{cl}(\tilde{\mathbf{X}},\tilde{\mathbf{X}}')
\end{equation}
where $\hat{\mathbf{Y}}$ denotes the pseudo-labels which are obtained as 
\begin{equation}\label{EQ:pseudo_label}
\hat{\mathbf{Y}}_i = \mathop{\mathrm{argmax}}\limits_{c}{\mathcal{P}(c|\mathbf{u}'_i)}
\end{equation}
where $\mathcal{P}(c|\mathbf{u}'_i)$ denotes the probability of class $c$ for the input data $\mathbf{u}'_i$.
Then, we calculate the attribution matrix via Eq.(\ref{EQ:LGAM2}) and conduct GMCL training with AttGMA module.

\section{Experiments}
In this section, we validate the effectiveness and benefits of the proposed GMCL on various graph learning tasks which include unsupervised learning, semi-supervised learning and transfer learning tasks. 
We implement our GMCL based on three main augmentation strategies, i.e., dropping, perturbation and mixup, which are denoted as 
\textbf{GMCL-D}, \textbf{GMCL-P} and \textbf{GMCL-M} respectively. 
All experiments are conducted on the computer with NVIDIA RTX A6000 GPU. 

\subsection{Unsupervised Learning}

\subsubsection{Experimental setup}
For the unsupervised learning tasks, we evaluate our GMCL on four biochemical molecules datasets, i.e., MUTAG~\cite{tudataset}, DD~\cite{tudataset}, PROTEINS~\cite{tudataset} and NCI1~\cite{tudataset}, and four social network datasets, i.e., IMDB-BINARY~\cite{tudataset}, COLLAB~\cite{tudataset}, REDDIT-B~\cite{tudataset} and REDDIT-M-5K~\cite{tudataset}.
The detailed information of these datasets are summarized in Table~\ref{tab:dataset-graph}.
\begin{table}[!ht]
    \centering
    \caption{Introduction of graph datasets for unsupervised learning tasks.}
    \label{tab:dataset-graph}
    \setlength{\tabcolsep}{1mm}{
    \begin{tabular}{c|c|c|c|c}
    \hline
    \hline
    \makecell{Bioinformatics \\ datasets} & MUTAG  & DD & PROTEINS & NCI1 \\
    \hline 
    Max \# vertices  & 28 & 5748 & 620 & 104  \\
    Mean \# vertices & 17.93 & 284.30 & 39.06 & 29.69  \\
    Mean \# edges & 19.79 & 715.65 & 72.82 & 64.29  \\
    \hline
    \# Graphs  & 188 & 1178 & 1113 & 4110  \\
    \# Classes & 2 & 2 & 2 & 2 \\
    \hline
    \hline
    \makecell{Social \\ datasets} & IMDB-B & REDDIT-B & REDDIT-M-5K & COLLAB\\
     \hline 
    Max \# vertices & 136 & 3783 & 3117 & 492\\
    Mean \# vertices & 19.77 & 429.61 & 523.41 & 74.49\\
    Mean \# edges & 4914.99 & 131.87 & 1228.35 & 4914.99 \\
    \hline
    \# Graphs  & 1,000 & 2000 & 500 & 497.80 \\
     \# Classes & 2 & 2 & 5 & 3 \\
  \hline
   \hline
    \end{tabular}}
\end{table}
We adopt the commonly used five-layer GIN~\cite{gin} as the learning backbone. 
The Encoder and MLP used in AttGMA module are 2-layer GIN and 2-layer full connected network respectively. 
The learning rate is set to $0.001$ and the maximum epoch is set to $500$. 
The hidden dimension is set to 128. 
The batch size is set to 128 for  GMCL-D, GMCL-P and 32 for GMCL-M. 
We use Adam~\cite{Adam} as the optimizer and all parameters in GMCL are optimized by minimizing the loss introduced in $\S$ 5.2.

\subsubsection{Comparison results}
To verify the effectiveness of the proposed methods, we compare our GMCLs against various alternative methods.
We first compare our GMCLs with three kernel-based methods including Graphlet Kernel (GL)~\cite{gk}, Weisfeiler-Lehman sub-tree kernel
(WL)~\cite{wlsubtree} and Deep Graph Kernel (DGK)~\cite{datasets}.
We also compare GMCLs with three unsupervised graph representation methods, i.e.,
node2vec~\cite{grover2016node2vec}, sub2vec~\cite{adhikari2018sub2vec} and graph2vec~\cite{narayanan2017graph2vec}.
In addition, we further compare our GMCLs with  some other recent graph contrastive learning methods which include InfoGraph~\cite{infograph},
GraphCL~\cite{graphcl}, GRACE~\cite{zhu2020deep}, JOAOv2~\cite{JOAO}, AD-GCL~\cite{ADGCL} and AutoGCL~\cite{autoGCL}. Note that, 
GraphCL and GRACE are GCL methods employing hybrid GDAs. JOAOv2 adopts the selective GDA in its learning process. 
AD-GCL and AutoGCL use the learnable GDA in their contrastive learning process. 
The results of GraphCL are obtained by using the default augmentation policy `random4' which selects two GDA methods randomly in each epoch. 
For GRACE~\cite{zhu2020deep}, we obtain its results by utilizing the source code provided by the authors. 
For the other comparison methods, the reported results are directly obtained from their published papers. 
For unsupervised learning tasks, the backbone network is trained with self-supervised contrastive loss to learn graph representations, which are then fed into a simple classifier for evaluation.
We conduct 10-fold cross validation on each dataset and present the average results of GMCLs in Table~\ref{tab:unsupervised}.

Table~\ref{tab:unsupervised} summarizes all comparison results. 
Overall, we can note that our GMCL can obtain the best performance on all datasets. 
In particular, we can note that compared with some other hybrid GDA methods (GraphCL~\cite{graphcl} and GRACE~\cite{zhu2020deep}) and 
selective GDA method JOAOv2~\cite{JOAO},
our proposed methods generally obtain the best performance on most datasets. 
Especially on dataset MUTAG, DD, IMDB-B and REDDIT-B, 
our GMCL methods can achieve almost 4\% improvement compared to these methods. 
These clearly demonstrate that our designed unified GMA methods perform more effectively on conducting unsupervised graph representation and learning tasks. 
In addition, our methods perform better than previous learnable GDA methods including AutoGCL~\cite{autoGCL} and AD-GCL~\cite{ADGCL} on most datasets.
Specifically, on IMDB-B and REDDIT-B datasets, our  GMCL methods obviously outperform AutoGCL~\cite{autoGCL}, 
which indicates that the proposed attribution based AttGMA module can better guide the preservation of label information in data augmentation process and thus lead to better label prediction results. 
\begin{table*}[!ht]
    \centering
    \renewcommand\arraystretch{1.2}
    \caption{Comparison results of unsupervised learning tasks on different datasets. The color of \textcolor{red}{red} and \textcolor{blue}{blue} denote the best and second performance respectively.}
    \label{tab:unsupervised}
    \begin{tabular}{c|c|c|c|c|c|c|c|c}
    \hline\hline
        Dataset & MUTAG & DD & PROTEINS & NCI1 & IMDB-B & REDDIT-B & REDDIT-M-5K & COLLAB \\ 
        \hline\hline
        GL~\cite{gk} & 81.66$\pm$2.11 & - & - & - & 65.87$\pm$0.98 & 77.34$\pm$0.18 & 41.01$\pm$0.17 & - \\ 
        WL~\cite{wlsubtree} & 80.72$\pm$3.00 & 72.92$\pm$0.56 & - & 80.01$\pm$0.50 & 72.30$\pm$3.44 & 68.82$\pm$0.41 & 46.06$\pm$0.21 & - \\
        DGK~\cite{datasets} & 87.44$\pm$2.72 & 73.30$\pm$0.82 & - & 80.31$\pm$0.46 & 66.96$\pm$0.56 & 78.04$\pm$0.39 & 41.27$\pm$0.18 & - \\ \hline
        node2vec~\cite{grover2016node2vec} & 72.63$\pm$10.20 & 57.49$\pm$3.57 & - & 54.89$\pm$1.61 & - & - & - & - \\
        sub2vec~\cite{adhikari2018sub2vec} & 61.05$\pm$15.80 & 53.03$\pm$5.55 & - & 52.84$\pm$1.47 & 55.26$\pm$1.54 & 71.48$\pm$0.41 & 36.68$\pm$0.42 & - \\
        graph2vec~\cite{narayanan2017graph2vec} & 83.15$\pm$9.25 & 73.30$\pm$2.05 & - & 73.22$\pm$1.81 & 71.10$\pm$0.54 & 75.78$\pm$1.03 & 47.86$\pm$0.26 & - \\ \hline
        InfoGraph~\cite{infograph} & 89.01$\pm$1.13 & 74.44$\pm$0.31 & 72.85$\pm$1.78 & 76.20$\pm$1.06 & 73.03$\pm$0.87 & 82.50$\pm$1.42 & 53.46$\pm$1.03 & 70.65$\pm$1.13 \\
        GraphCL~\cite{graphcl} & 86.80$\pm$1.34 & \textcolor{blue}{78.62$\pm$0.40} &  74.39$\pm$0.45 & 77.87$\pm$0.41 & 71.14$\pm$0.44 & 89.53$\pm$0.84 & 55.99$\pm$0.28 & 71.36$\pm$1.15 \\
        GRACE~\cite{zhu2020deep} & 90.65$\pm$0.81 & 78.15$\pm$0.77 & 73.07$\pm$0.52 & 76.16$\pm$0.42 & 71.35$\pm$0.46 & 89.82$\pm$0.61 & 55.71$\pm$0.83 & 71.47$\pm$1.59 \\
        JOAOv2~\cite{JOAO} & 87.67$\pm$0.79 & 71.25$\pm$0.85 & 66.91$\pm$1.75 & 72.99$\pm$0.75 & 71.60$\pm$0.86 & 78.35$\pm$1.38 & 45.57$\pm$2.86 & 70.40$\pm$2.21 \\ 
        AD-GCL~\cite{ADGCL} & 89.42$\pm$3.29 & 73.59$\pm$0.65 & 74.49$\pm$0.52 & 69.67$\pm$0.51 & 71.57$\pm$1.01 & 85.52$\pm$0.79 & 53.00$\pm$0.82 & 73.32$\pm$0.61 \\ 
        AutoGCL~\cite{autoGCL} & 88.64$\pm$1.08 & 75.80$\pm$0.36 & \textcolor{red}{77.57$\pm$0.60} & 82.00$\pm$0.29 & 73.30$\pm$0.40 & 88.58$\pm$1.49 & \textcolor{blue}{56.75$\pm$0.18} & 70.12$\pm$0.68 \\ \hline
        GMCL-D & \textcolor{red}{94.09$\pm$6.28} & \textcolor{red}{78.78$\pm$3.38} & 76.92$\pm$3.12 &\textcolor{red}{82.99$\pm$2.06} & \textcolor{blue}{75.20$\pm$4.04} & 91.85$\pm$2.38 & 56.63$\pm$1.86 & 74.21$\pm$1.73 \\ 
        GMCL-P & \textcolor{blue}{93.10$\pm$5.33} & 77.59$\pm$3.68 & 76.82$\pm$2.57 & 81.99$\pm$2.18 & \textcolor{red}{75.60$\pm$3.17} & \textcolor{blue}{92.15$\pm$1.60} & 56.35$\pm$2.31 & \textcolor{blue}{74.76$\pm$1.37} \\ 
        GMCL-M & 92.54$\pm$6.85 & 78.36$\pm$2.41 & \textcolor{blue}{77.44$\pm$3.83} & \textcolor{blue}{82.75$\pm$1.70} & 75.10$\pm$4.16 & \textcolor{red}{92.50$\pm$1.86} & \textcolor{red}{56.81$\pm$2.50} & \textcolor{red}{75.88$\pm$1.18} \\ 
        \hline\hline
    \end{tabular}
\end{table*}

\subsection{Semi-supervised Learning}
\subsubsection{Experimental setup}
Following the previous work~\cite{autoGCL}, we utilize a 10-fold cross-validation approach for the semi-supervised learning task.
Each fold consists of 80\% of the total data as unlabeled data, 10\% as labeled training data and the remaining 10\% as testing data. 
We use 3-layer ResGCN~\cite{resgcn} as the GNN backbone for graph feature extraction and the classifier consists of one linear layer and one softmax layer. 
The Encoder and MLP used in the AttGMA module employ a 2-layer GIN and 2-layer full connected network respectively.
The learning rate is set to 0.001.
The number of hidden units and batch size are set to 128 and 64 respectively.
For the projection head, we employ the simple `sum' function. 
The value of hyperparameter $\alpha$ is estimated based on the validation performance. 

\subsubsection{Comparison results}

We compare our GMCLs with seven GCL methods including GCA~\cite{GCA}, Infomax~\cite{DGI}, GraphCL~\cite{graphcl}, GRACE~\cite{zhu2020deep}, JOAOv2~\cite{JOAO}, AD-GCL~\cite{ADGCL} and AutoGCL~\cite{autoGCL}.
For AD-GCL~\cite{ADGCL}, the reported results are obtained by running its source codes. 
The results for the other comparison methods on MUTAG dataset and the performance of JOAOv2~\cite{JOAO} on IMDB-B dataset are acquired by running the codes provided by the authors. 
The remaining results are directly obtained from previously published works~\cite{autoGCL,JOAO}. 
Table~\ref{tab:semisupervised} summarizes the experimental results on semi-supervised graph classification tasks. 
It can be observed that our methods show superior performance on most of the datasets. 
Specifically, comparing with GraphCL~\cite{graphcl}, the proposed GMCL-D, GMCL-P and GMCL-M 
respectively achieve average improvements of 3.4\%, 3.8\%, and 3.9\% on all datasets.
This demonstrates the benefits of the proposed adaptive GMA in guiding effective contrastive learning. 
Additionally, when compared to some other hybrid and selective GDA methods (GraphCL~\cite{graphcl}, GRACE~\cite{zhu2020deep} and JOAO~\cite{JOAO}), 
our GMCLs exhibit better performance on most datasets, which further validates the advantages of our GMA-based contrastive learning strategy. 
Finally, 
our GMCLs outperform previous learnable GDA-based methods such as AD-GCL~\cite{ADGCL} and AutoGCL~\cite{autoGCL}. 
It clearly demonstrates the effectiveness of the proposed attribution based AttGMA module which encourages to capture the label-invariant information for graph data representation and thus can improve the final semi-supervised classification performance. 

\begin{table*}[!ht]
    \centering
    \renewcommand\arraystretch{1.2}
    \caption{Comparison results of semi-supervised learning tasks on different datasets. The color of \textcolor{red}{red} and \textcolor{blue}{blue} denote the best and second performance respectively.}
    \label{tab:semisupervised}
    \begin{tabular}{c|c|c|c|c|c|c|c|c}
    \hline\hline
        Dataset & MUTAG & DD & PROTEINS & NCI1 & IMDB-B & REDDIT-B & REDDIT-M-5K & COLLAB \\ \hline
        Full Data & 93.07$\pm$3.41 & 74.36$\pm$5.86 & 69.72$\pm$6.71 & 75.16$\pm$2.07 & 64.80$\pm$4.92 & 76.75$\pm$5.60 & 49.71$\pm$3.20 & 74.34$\pm$2.00 \\ \hline
        GCA~\cite{GCA} & 80.79$\pm$10.51 & 76.74$\pm$4.09 & 73.85$\pm$5.56 & 68.73$\pm$2.36 & 73.70$\pm$4.88 & 77.15$\pm$6.96 & 32.95$\pm$10.89 & 74.32$\pm$2.30 \\ 
        Infomax~\cite{DGI} & - & 75.78$\pm$0.34 & 72.27$\pm$0.40 & 74.86$\pm$0.26 & - & 88.66$\pm$0.95 & 53.61$\pm$0.31 & 73.75$\pm$0.29 \\ 
        GraphCL~\cite{graphcl} & 85.60$\pm$6.85 & 76.65$\pm$5.12 & 74.21$\pm$4.50 & 73.16$\pm$2.90 & 68.10$\pm$5.15 & 78.05$\pm$2.65 & 48.09$\pm$1.74 & 75.50$\pm$2.15 \\ 
        GRACE~\cite{zhu2020deep}   & 84.41$\pm$5.57 & 76.82$\pm$4.96 & 73.80$\pm$2.95 & 74.09$\pm$3.28 & 68.21$\pm$5.28 & 79.33$\pm$3.28 & 49.96$\pm$1.82 & 75.32$\pm$2.56 \\
        JOAOv2~\cite{JOAO} & 89.67$\pm$5.24 & 75.81$\pm$0.73 & 73.31$\pm$0.48 & 74.86$\pm$0.39 & 68.30$\pm$4.08 & 88.79$\pm$0.65 & 52.71$\pm$0.28 & 75.53$\pm$0.18 \\ 
        AD-GCL~\cite{ADGCL} & 90.28$\pm$4.01 & 77.91$\pm$0.73 & 73.96$\pm$0.47 & 75.18$\pm$0.31 & 72.48$\pm$1.15 & 90.10$\pm$0.15 & 53.49$\pm$0.28 & 75.82$\pm$0.26 \\ 
        AutoGCL~\cite{autoGCL} & 88.30$\pm$5.48 & 77.50$\pm$4.41 & 75.65$\pm$2.40 & 73.75$\pm$2.25 & 71.90$\pm$2.88 & 79.80$\pm$3.47 & 49.91$\pm$2.70 & \textcolor{red}{77.16$\pm$1.48} \\ \hline
        GMCL-D & 90.40$\pm$5.40 & 79.50$\pm$4.30 & 77.20$\pm$2.50 & 76.90$\pm$1.70 & \textcolor{red}{74.30$\pm$4.30} & \textcolor{blue}{91.00$\pm$2.61} & \textcolor{red}{55.50$\pm$1.30} & 76.40$\pm$1.40 \\ 
        GMCL-P & \textcolor{blue}{92.50$\pm$4.50} & \textcolor{blue}{80.00$\pm$4.10} & \textcolor{red}{77.90$\pm$1.80} & \textcolor{red}{77.50$\pm$1.70} & \textcolor{blue}{73.80$\pm$3.90} & 90.90$\pm$2.40 & \textcolor{blue}{55.50$\pm$2.10} & 76.44$\pm$1.67 \\
        GMCL-M & \textcolor{red}{93.10$\pm$6.30} & \textcolor{red}{80.50$\pm$3.00} & \textcolor{blue}{77.80$\pm$2.50} & \textcolor{blue}{77.00$\pm$2.00} & 73.80$\pm$4.75 & \textcolor{red}{91.20$\pm$3.00} & 55.30$\pm$1.80 & \textcolor{blue}{76.70$\pm$1.80} \\ 
        \hline\hline
    \end{tabular}
\end{table*}

\subsection{Transfer Learning}
\subsubsection{Experimental setup}
We conduct evaluation on some molecular datasets to validate the transfer ability of the proposed GMCLs. 
Specifically, we evaluate the performance of the GNN trained by using the proposed GMCL on predicting the properties of chemical molecules and functions of biological proteins. 
Following the previous work~\cite{autoGCL}, we employ the same experimental setup, i.e.,   first utilizing self-supervised learning for pre-training GNN on the larger molecular dataset ChEMBL~\cite{gaulton2012chembl} and then conducting fine-tuning on six smaller molecular datasets to evaluate the out-of-distribution performance. 
The data splittings of these six datasets are similar to previous related works~\cite{graphcl,autoGCL}.
For transfer learning task, we employ the widely used 5-layer GIN~\cite{gin} as the backbone and use the `mean' function as the projection head. 
The learning rate is set to 0.001 in both pre-training and fine-tuning stages.
The final embedding dimension and the number of hidden units are set to 300 and 128 respectively. 
For pre-training, the batch size is set to 128. 
For fine-tuning, it is set to 32 on smaller datasets. 
The entire model is optimized by using Adam~\cite{Adam} algorithm. 
We pre-train the GNN backbone via our GMCL framework for 30 epochs and fine-tune it on the other datasets for 500 epochs. 

\subsubsection{Comparison results}
For transfer learning task,
we compare our GMCLs with nine GCL methods including 
Infomax~\cite{DGI}, EdgePred~\cite{hu2020strategies}, AttrMasking~\cite{hu2020strategies}, ContextPred~\cite{hu2020strategies},
GraphCL~\cite{graphcl}, GRACE~\cite{zhu2020deep}, JOAOv2~\cite{JOAO}, AD-GCL~\cite{ADGCL} and AutoGCL~\cite{autoGCL}.
The performance of these methods are directly obtained from previously published works~\cite{autoGCL,JOAO}. 
We report the average results of ten different data splits for all comparison methods and GMCLs and summarize it in Table~\ref{tab:transfer}.
Here, we can note that, 
(1) comparing with model without any pre-training process, our GMCLs can achieve significant improvement, particularly on  ClinTox, MUV, BACE and BBBP datasets. 
This indicates the desired benefit of the proposed self-supervised learning method for pre-training GNN models.  
(2) Our GMCLs outperform some state-of-the-art GCL methods with randomly generated GDA including Infomax~\cite{DGI}, EdgePred~\cite{hu2020strategies}, AttrMasking~\cite{hu2020strategies}, ContextPred~\cite{hu2020strategies} and the baseline model GraphCL~\cite{graphcl} on most of the datasets. 
It further demonstrates the more effective of the proposed message contrastive learning when compared with traditional GCL methods. 
(3) GMCLs demonstrate better performance than traditional GCL methods with learnable and selective GDAs (JOAOv2~\cite{JOAO}, AD-GCL~\cite{ADGCL} and AutoGCL~\cite{autoGCL}) on most datasets. This further indicates the advantage of the proposed AttGMA module on guiding adaptive contrastive learning. 

\begin{table*}[!ht]
    \centering
    \renewcommand\arraystretch{1.2}
    \caption{Comparison results on transfer learning tasks on different datasets. The color of \textcolor{red}{red} and \textcolor{blue}{blue} denote the best and second performance respectively.}
    \label{tab:transfer}
    \begin{tabular}{c|c|c|c|c|c|c}
    \hline
    \hline
        Dataset &  ToxCast & ClinTox & MUV & HIV & BACE & BBBP \\ \hline
        No Pretrain &  63.4$\pm$0.6 & 58.0$\pm$4.4 & 71.8$\pm$2.5 & 75.3$\pm$1.9 & 70.1$\pm$5.4 & 65.8$\pm$4.5 \\ \hline
        Infomax~\cite{DGI} &  62.7$\pm$0.4 & 69.9$\pm$3.0 & 75.3$\pm$2.5 & 76.0$\pm$0.7 & 75.9$\pm$1.6 & 68.8$\pm$0.8 \\
        EdgePred~\cite{hu2020strategies} & 64.1$\pm$0.6 & 64.1$\pm$3.7 & 74.1$\pm$2.1 & 76.3$\pm$1.0 & 79.9$\pm$0.9 & 67.3$\pm$2.4 \\
        AttrMasking~\cite{hu2020strategies} & \textcolor{blue}{64.2$\pm$0.5} & 71.8$\pm$4.1 & 74.7$\pm$1.4 & 77.2$\pm$1.1 & 79.3$\pm$1.6 & 64.3$\pm$2.8 \\
        ContextPred~\cite{hu2020strategies} &  63.9$\pm$0.6 & 65.9$\pm$3.8 & 75.8$\pm$1.7 & 77.3$\pm$1.0 & 79.6$\pm$1.2 & 68.0$\pm$2.0 \\\hline
        GraphCL~\cite{graphcl} &  62.40$\pm$0.57 & 75.99$\pm$2.65 & 69.80$\pm$2.66 & 78.47$\pm$1.22 & 75.38$\pm$1.44 & 69.68$\pm$0.67 \\
        GRACE~\cite{zhu2020deep}   &  61.17$\pm$0.53 & 76.39$\pm$2.31 & 68.87$\pm$2.38 & 77.91$\pm$1.29 & 75.42$\pm$1.31 & 69.57$\pm$0.78 \\
        JOAOv2~\cite{JOAO} &  63.16$\pm$0.45 & 80.97$\pm$1.64 & 73.67$\pm$1.00 & 77.51$\pm$1.17 & 75.49$\pm$1.27 & 71.39$\pm$0.92 \\
        AD-GCL~\cite{addrop} &  63.07$\pm$0.72 & 79.78$\pm$3.52 & 72.30$\pm$1.61 & 78.28$\pm$0.97 & 78.51$\pm$0.80 &  70.01$\pm$1.07 \\
        AutoGCL~\cite{autoGCL} &  63.47$\pm$0.38 & 80.99$\pm$3.38 & 75.83$\pm$1.30 & 78.35$\pm$0.64 & \textcolor{red}{83.26$\pm$1.13} & 73.36$\pm$0.77 \\ \hline
        GMCL-D &  64.16$\pm$0.44 & 81.16$\pm$3.45 & \textcolor{red}{78.03$\pm$1.56} & \textcolor{red}{79.88$\pm$0.76} & \textcolor{blue}{82.78$\pm$0.70} & 73.49$\pm$0.73 \\ 
        GMCL-P &  \textcolor{red}{64.21$\pm$0.33} & \textcolor{blue}{81.62$\pm$3.78} & 77.48$\pm$0.51 & \textcolor{blue}{79.47$\pm$0.80} & 82.66$\pm$1.19 & \textcolor{blue}{73.61$\pm$1.06} \\
        GMCL-M & 64.12$\pm$0.31 & \textcolor{red}{83.37$\pm$3.83} & \textcolor{blue}{77.61$\pm$1.28} & 79.28$\pm$0.89 & 82.37$\pm$1.16 & \textcolor{red}{73.80$\pm$1.21} \\
    \hline
    \hline
    \end{tabular}
\end{table*}

\subsection{Generalizability  Evaluation}

As discussed in $\S$ 4, 
the proposed GMA provides a general data augmentation for graph data. 
To demonstrate the generalizability of GMA,
we evaluate our methods on four kinds of datasets including two bioinformatics datasets (MUTAG~\cite{tudataset} and DD~\cite{tudataset}), two social datasets (IMDB-B~\cite{tudataset} and REDD-B~\cite{tudataset}), two molecular datasets (MUV~\cite{gaulton2012chembl} and HIV~\cite{gaulton2012chembl}) and two citation datasets (Cora~\cite{datasets} and Citeseer~\cite{datasets}).
We report the results of unsupervised learning on bioinformatics and social datasets, transfer learning on molecular datasets and semi-supervised node classification on citation datasets. 
For citation datasets, 
we use the standard data partition used in many previous works~\cite{gcn,gat}.
All methods use a 2-layer GCN with 32 hidden units as the baseline for node classification tasks. 
The parameters are optimized by using both contrastive loss and classification loss for semi-supervised tasks. 
We compare the effectiveness of the proposed 
GMAs with nine typical GDA methods including five dropping-based GDAs, i.e., node dropping, edge dropping, node attribute dropping, edge attribute dropping and subgraph sampling, and four perturbation-based GDAs, i.e., node perturbation, edge perturbation, node feature perturbation and edge attribute perturbation. 
To ensure fairness, we employ the same attribution-guided learnable strategy for these compared GDAs which is similar to the proposed AttGMA. 
Specifically, for GDA on node set, node attributes and subgraph type GDA, we first use GNN encoder to compute the probability matrix and then adopt
Eq.(\ref{EQ:LGAM2}) and Eq.(\ref{EQ:LGAM3})) to obtain the corresponding indicator matrix.
For GDA on edge set and edge attributes, we first use GIN encoder followed by MLP to compute the probability matrix and then utilize
Eq.(\ref{EQ:LGAM2}) and Eq.(\ref{EQ:LGAM3})) to obtain the corresponding indicator matrix.
For GDA on node set, edge set, node attributes and edge attributes, we conduct the augmentation operations as 
Eqs.(\ref{EQ:GDAN},~\ref{EQ:GDAEF}) respectively.
Note that, for the molecular dataset, the perturbation operation on edge is equivalent to the perturbation on edge attributes.
Thus, we do not report the results of edge perturbation methods on this dataset. 
Table~\ref{tab:message_as} summarizes the comparison results on all datasets. 
Since bioinformatics, social and citation datasets lack edge attributes, the results of GDA with edge attribute perturbation are not reported in Table~\ref{tab:message_as}.
From Table~\ref{tab:message_as}, 
we can observe that different datasets exhibit preferences for different GDA methods. 
For example, 
both node dropping and edge dropping are more suitable for bioinformatics and social datasets when compared to perturbing node features. 
In contrast, our GMA methods demonstrate consistently superior performance on multiple datasets. 
This further shows the generalizability of the proposed GMA on various graph learning scenarios and tasks. 

\begin{table*}[!ht]
    \setlength\tabcolsep{4pt}
    \centering
    \renewcommand\arraystretch{1.2}
    \caption{Comparison results of various GDA methods and our GMAs on four different kinds of datasets. The color of \textcolor{red}{red} and \textcolor{blue}{blue} denote the best and second performance respectively.}
    \label{tab:message_as}
    \begin{tabular}{c|c|c|c|c|c|c|c|c|c}
    \hline \hline
        \multirow{2}{*}{Type} & \multirow{2}{*}{Method} & \multicolumn{2}{c|}{Bioinformatics datasets}&\multicolumn{2}{c|}{Social datasets}& \multicolumn{2}{c|}{Molecular datasets}& \multicolumn{2}{c}{Citation datasets} \\ \cline{3-10}
        & & MUTAG & DD & IMDB-B & REDDIT-B & MUV & HIV & Cora & Citeseer\\ \hline
        \multirow{9}{*}{GDA} & Node drop & 91.49$\pm$5.87 & 77.33$\pm$3.22 & 73.60$\pm$4.39 & 90.35$\pm$2.28 & 75.88$\pm$1.41 & 78.26$\pm$0.73  & 83.44$\pm$0.39 & 73.53$\pm$0.37 \\ 
        & Node perturb  & 90.96$\pm$4.26 & 76.40$\pm$4.21 & 72.30$\pm$4.84 & 91.25$\pm$1.49 & 76.49$\pm$1.69 & 78.19$\pm$0.55 & 82.88$\pm$0.21 & 73.62$\pm$0.39 \\ \cline{2-10}
        & Edge drop & 91.46$\pm$6.12 & 78.18$\pm$2.80 & 74.20$\pm$4.24 & 89.25$\pm$2.32 & 76.17$\pm$1.48 & 78.94$\pm$0.94 & 83.20$\pm$0.40 & 72.86$\pm$0.22 \\ 
        & Edge perturb  & 90.38$\pm$5.28 & 77.33$\pm$5.61 & 74.60$\pm$2.69 & 91.45$\pm$1.74 & - & - & 81.72$\pm$0.59 & 73.43$\pm$0.37 \\ \cline{2-10}
        & Node attribute drop  & 91.46$\pm$4.28 & 77.84$\pm$2.13 & 74.20$\pm$4.14 & 90.35$\pm$2.28 & 76.93$\pm$1.45 & 78.61$\pm$0.29 & 83.53$\pm$0.33 & 73.57$\pm$0.27 \\ 
        & Node attribute perturb & 90.96$\pm$3.37 & 75.89$\pm$2.74 & 73.70$\pm$3.41 & 90.25$\pm$2.25 & 75.79$\pm$1.33 & 78.24$\pm$0.84 & 83.45$\pm$0.41 & 73.50$\pm$0.36 \\ \cline{2-10}
        & Edge attribute drop  & - & - & - & - & 77.13$\pm$1.67 & 79.14$\pm$0.37 & - & - \\ 
        & Edge attribute perturb         & - & - & - & - & 76.04$\pm$1.47 & 77.91$\pm$0.99 & - & - \\ \cline{2-10}
        & Subgraph  & 90.00$\pm$5.98 & 76.99$\pm$4.43 & 74.30$\pm$3.72 & 90.30$\pm$2.64 & 77.18$\pm$2.11 & 79.29$\pm$0.68 & 82.66$\pm$0.20 & 73.20$\pm$0.46 \\ \hline
        \multirow{3}{*}{GMA} & Message drop & 92.08$\pm$4.84 & \textcolor{red}{78.94$\pm$2.99} & \textcolor{blue}{75.20$\pm$4.04} & 91.85$\pm$2.38 & \textcolor{red}{78.03$\pm$1.56} & \textcolor{red}{79.88$\pm$0.76} & \textcolor{blue}{84.30$\pm$0.62} & \textcolor{red}{74.48$\pm$0.37}  \\ 
        & Message perturb & \textcolor{red}{93.10$\pm$5.33} & 77.59$\pm$3.68 & \textcolor{red}{75.60$\pm$3.17} & \textcolor{blue}{92.15$\pm$1.60} & 77.48$\pm$0.51 & \textcolor{blue}{79.47$\pm$0.80} & 83.97$\pm$0.21 & \textcolor{blue}{74.48$\pm$0.53} \\ 
        & Message mixup & \textcolor{blue}{92.54$\pm$6.85} & \textcolor{blue}{78.36$\pm$2.41} & 75.10$\pm$4.16 & \textcolor{red}{92.50$\pm$1.86} & \textcolor{blue}{77.61$\pm$1.28} & 79.28$\pm$0.89 & \textcolor{red}{84.88$\pm$0.27} & 74.46$\pm$0.38\\ \hline \hline
    \end{tabular}
\end{table*}
\begin{figure*}[!htpb]
\centering
\includegraphics[width=0.95\textwidth]{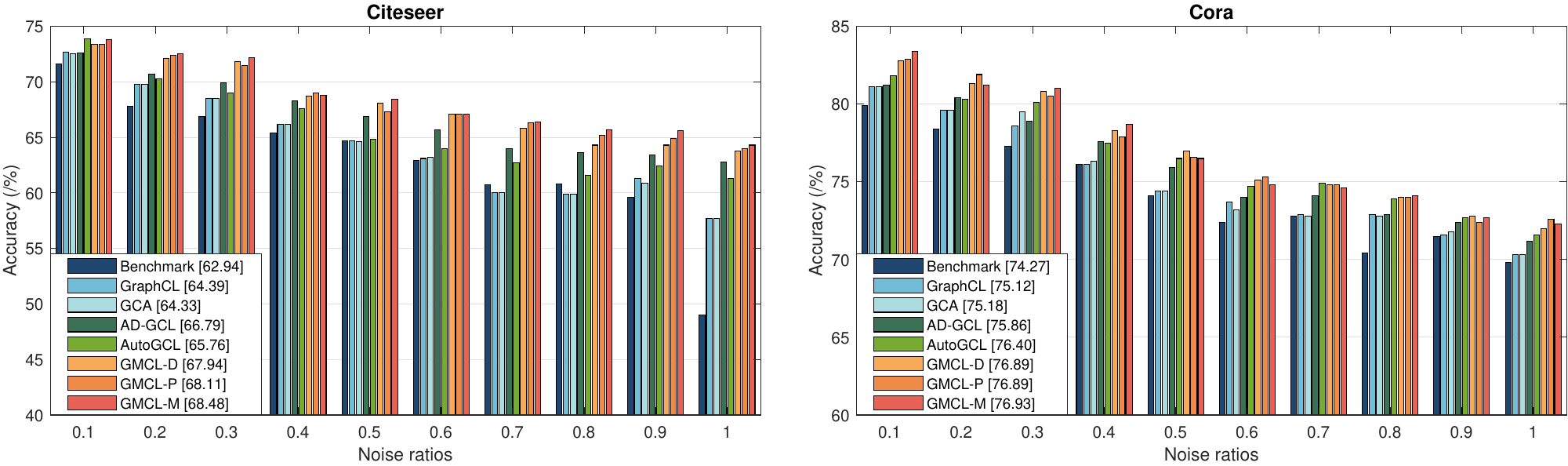}
  \caption{The comparison performance of GMCL on two citation datasets with different ratios of random attack noises.}
\label{fig:noise}
\end{figure*}
\subsection{Robustness Evaluation}

To verify the robustness of the proposed GMCLs,
we conduct evaluation on the commonly used citation dataset Cora~\cite{sen2008collective} and Citeseer~\cite{sen2008collective} and apply the random attack method~\cite{deeprobust} to introduce structural noises.
We utilize 10 different ratios of structural noises varying from 0.1 to 1 with step size 0.1.
Similar to many previous works~\cite{gcn,gat},
we use standard data splitting for citation datasets.
We compare GMCLs with the baseline method GraphCL~\cite{graphcl} and three other graph contrastive methods including GCA~\cite{GCA}, 
AD-GCL~\cite{ADGCL} and AutoGCL~\cite{autoGCL}.
All methods use a 2-layer GCN with 32 hidden units as the baseline. 
The parameters in all GCL frameworks are optimized by using both contrastive loss and classification loss simultaneously. 
GraphCL~\cite{graphcl} uses the default augmentation policy `random4' with 50\% dropout ratio.
Additionally, for AutoGCL~\cite{autoGCL}, we use a similarity loss to maintain consistency with the original method proposed in its paper. 
All compared models are trained by using the Adam optimizer~\cite{Adam}. 
\begin{figure*}[!htpb]
\centering
\includegraphics[width=0.95\textwidth]{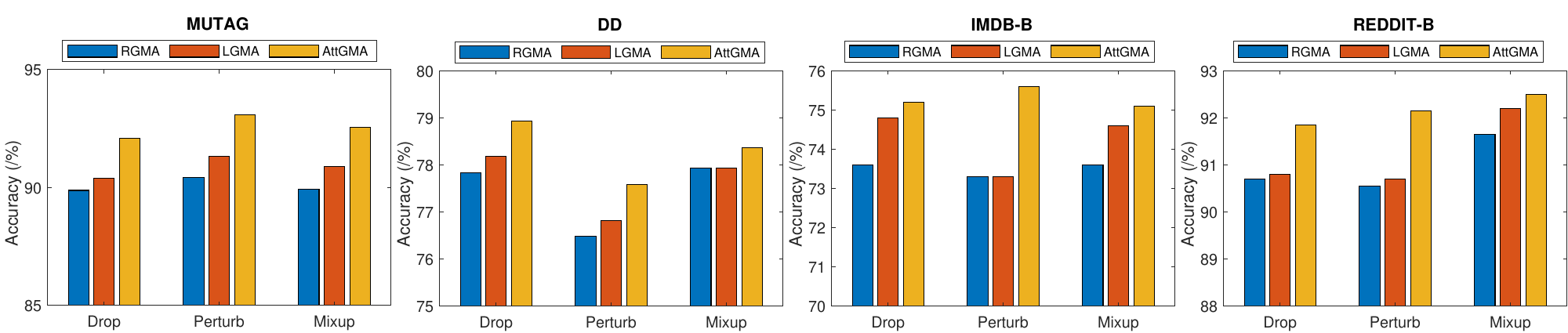}
  \caption{Comparison results of GMCL with RGMA, LGMA and AttGMA respectively on four different datasets.}
\label{fig:attribution_as}
\end{figure*}
Fig.~\ref{fig:noise} demonstrates the comparative results on two citation datasets with various ratios of structural noises. 
We can observe that all GCL frameworks are capable of strengthening model's robustness. 
Also, the incorporation of learnable GDAs can further improve the stability of models in the presence of structural noises. 
GCA, AutoGCL and AD-GCL that utilize learnable node dropping and edge dropping based GDA strategies can outperform  GraphCL which employs non-learnable GDA strategies on three datasets. 
Note that our proposed methods demonstrate superior performance on all datasets which demonstrates the more robustness of the proposed unified GMA methods and attribution-guided adaptive augmentor approach. 

\subsection{Ablation Analysis}
In this section, 
we perform ablation experiments to validate the effectiveness of our proposed  AttGMA module on two bioinformatics datasets (MUTAG~\cite{tudataset} and DD~\cite{tudataset})
and two social network datasets (IMDB-B~\cite{tudataset} and REDDIT-B~\cite{tudataset}). 
Here, we implement a variant of the proposed method (denoted as LGMA)  by 
directly sampling message elements based on $\mathbf{P}$ in Eq.(\ref{EQ:LGAM3}). 
Fig.~\ref{fig:attribution_as} shows the comparisons of the proposed AttGMA with LGMA and randomly GMA (denotes RGMA). 
%
Here, we can observe that both learnable augmentor LGMA and AttGMA generally bring more benefits to graph contrastive learning tasks when compared to random augmentor (RGMA). 
Moreover, 
the proposed AttGMA generally outperforms traditional learnable approach (LGMA), which further demonstrates the more effectiveness of the proposed attribution-guided augmentor in graph contrastive learning. 

\section{Conclusion}

In this paper, we re-think graph data augmentation in graph contrastive learning and introduce a novel universal data augmentation method, termed Graph Message Augmentation (GMA), for graph data representation learning. 
GMA provides a general formulation for graph data augmentation. 
It can unify many traditional GDA methods and provide a unified understanding for many previous GDAs. 
Also, GMA provides a natural way to implement mixup operation on graphs. 
Based on GMA, we then propose a general Graph Message Contrastive Learning (GMCL) for graph self-supervised learning problems.  
An attribution-guided augmentation strategy is further introduced to enhance the effectiveness and robustness of GMCLs. 
Experimental results on various graph data learning tasks across many different datasets demonstrate the effectiveness, generalizability and robustness of the proposed GMCL approach.

\bibliographystyle{ieeetr}
\bibliography{REFERENCES}

\end{document}